\documentclass[10pt,conference]{IEEEtran}

\usepackage[utf8]{inputenc}
\usepackage{graphicx}
\usepackage{amsmath,amssymb}
\usepackage{url}
\usepackage{subfigure}
\usepackage[english]{babel}
\usepackage{caption}
\usepackage{booktabs}
\usepackage[dvipsnames]{xcolor}
\usepackage{hyperref}

\makeatletter
\def\endthebibliography{%
  \def\@noitemerr{\@latex@warning{Empty `thebibliography' environment}}%
  \endlist
}
\makeatother

\title{Can we learn gradients by \\ Hamiltonian Neural Networks?}
\author{Aleksandr Timofeev, Andrei Afonin, Yehao Liu\\
  \textit{EPFL, Switzerland}}

\begin{document}
  
\maketitle

\begin{abstract}
In this work, we propose a meta-learner based on ODE neural networks that learns gradients. This approach makes the optimizer is more flexible inducing an automatic inductive bias to the given task. Using the simplest Hamiltonian Neural Network we demonstrate that our method outperforms a meta-learner based on LSTM for an artificial task and the MNIST dataset with ReLU activations in the optimizee. Furthermore, it also surpasses the classic optimization methods for the artificial task and achieve comparable results for MNIST.
\end{abstract}

\section{Introduction}
Nowadays, deep learning is successfully applied in many fields \cite{abiodun2018state}, especially in image recognition and natural language processing. This success is due to approximation power of the neural networks \cite{hornik1989multilayer} and effective application of the manually designed first-order optimization methods \cite{kingma2014adam, duchi2011adaptive, nesterov1983method}. But the design of an optimization algorithm can be considered as a learning problem that hopefully will give better convergence results due to adjustment to a particular task. 

The No Free Lunch Theorem \cite{wolpert1997no} suggests that there is no universally best learner and restricting the hypothesis class by introducing our prior knowledge about the task we are solving is the only way we can improve the state of affairs. This motivates the use of the learned optimizer for the given task and the use of different regularization methods. For instance, the Heavy Ball method \cite{polyak1964some} considers the gradient descent procedure as a sliding of a heavy ball on the surface of the loss function, which results in faster convergence. More generally, one can consider the gradient descent procedure as a movement of some object on the surface of the loss function under different forces: potential, dissipative (friction) and other external forces. Such a physical process can be described by port-Hamiltonian system of equations \cite{van2006port}. Work \cite{massaroli2019porthamiltonian} considers the optimization process as the evolution of a port-Hamiltonian system meaning that the parameters of the neural network are the solutions of the port-Hamiltonian system of equations. The results show that this framework helps to overcome the problem of getting stuck at saddle points which motivates its use for the non–convex, high–dimensional neural networks. In this work, we propose to learn the optimizer and impose the physical laws governed by the port-Hamiltonian system of equations into the optimization algorithm to provide implicit bias which acts as regularization and helps to find the better generalization optimums. We impose physical structure by learning the gradients of the parameters: gradients are the solutions of the port-Hamiltonian system, thus their dynamics is governed by the physical laws, that are going to be learned.

To summarize, we propose a new framework based on Hamiltonian Neural Networks which is used to learn and improve gradients for   the gradient descent step. Our experiments on an artificial task and MNIST dataset demonstrate that our method is able to outperform many basic optimizers and achieve comparable performance to the previous LSTM-based one. Furthermore, we explore how methods can be transferred to other architectures with different hyper-parameters, e.g. activation functions. To this end, we train HNN-based optimizer for a small neural network with the sigmoid activation on MNIST dataset and then train the same network but with the ReLU activation using the already trained optimizer. The results show that our method is transferable in this case unlike the LSTM-based optimizer. The implementation is uploaded to GitHub: \textcolor{blue}{\url{https://github.com/AfoninAndrei/OPT-ML}}.

\section{Methodology}

Our work is mainly based on \cite{andrychowicz2016learning}. Given the some learning task and an objective function $F(\theta)$ defined over some domain $\theta \in \Theta$, the goal is to find the global minimum $\theta^{*} = \arg \min_{\theta \in \Theta} F(\theta)$. Usual approach in the deep learning is to use the gradient based update rule: 
\begin{gather} \label{GD update}
    \theta_{t+1} = \theta_{t} - \gamma_{t} \nabla F(\theta_{t}),
\end{gather}
where $\nabla F(\theta_{t})$ is the gradient of the objective $F(\theta)$ at the point $\theta_{t}$ and $\gamma_{t}$ is the step size. Similar to \cite{andrychowicz2016learning} we propose to use the following update rule instead of Equation \eqref{GD update}:
\begin{gather*} \label{eq: update}
    \theta_{t+1} = \theta_{t} - \gamma_{t} f_{\phi}(\nabla F(\theta_{t}), \dot{\nabla F(\theta_{t})}),
\end{gather*}
where $f_{\phi}(\nabla F(\theta_{t}), \dot{\nabla F(\theta_{t})})$ is the output of the optimizer neural network $f_{\phi}$ with the parameters $\phi$, inputs $\nabla F(\theta_{t})$ and $\dot{\nabla F(\theta_{t})}$ where the last is the time derivative of the gradient at the point $\theta_{t}$. We propose to think of the gradient $\nabla F$ as a physical system with the continuous evolution governed by some laws. Physical structure is encoded into the architecture of the neural network $f_{\phi}$ which is motivated by \cite{zhong2020symplectic, zhong2020dissipative}:
\begin{equation} \label{eq: port-Hamilt}
   \begin{bmatrix} 
        \dot{\text{q}} \\ 
        \dot{\text{p}}
    \end{bmatrix} = 
    \left(\begin{bmatrix} 
    0 & \text{E}\\ -\text{E} & 0\end{bmatrix}-\text{D}_{\phi_2}(\text{q})\right) 
    \begin{bmatrix} 
        \frac{\partial H_{\phi_1}(\text{q},\text{p})}{\partial \text{q}} \\ 
        \frac{\partial H_{\phi_1}(\text{q},\text{p})}{\partial \text{p}}
    \end{bmatrix} +
    \begin{bmatrix} 
        0 \\ 
        \text{G}_{\phi_3}(\text{q})
    \end{bmatrix},
\end{equation}
\begin{equation*} \label{eq: Hamiltonian}
    H_{\phi_1}(\text{q},\text{p}) = \frac{1}{2}\text{p}^T \text{M}^{-1}_{\phi_{1,1}}(\text{q}) \text{p} + \text{V}_{\phi_{1,2}}(\text{q}),
\end{equation*}
where $\text{M}$ is an inertia matrix, $\text{q}$ and $\text{p}$ are the generalized coordinate and impulse of the physical system correspondingly. Their derivatives by time are $\dot{\text{q}}$ and $\dot{\text{p}}$ and notice that $\text{p} = \text{M}\dot{\text{q}}$. $H$ is a Hamiltonian, $\text{V}$ is a potential, $\text{D}$ is a dissipative term and $\text{G}$ are external forces which are dependant only on the coordinate $\text{q}$ and affect only the impulse $\text{p}$. This form of $\text{G}$ is applicable for many physical systems. Thus, inertia matrix $\text{M}$, potential term $\text{V}$, dissipative matrix $\text{D}$ and external forces $\text{G}$ are approximated by the neural networks $\text{M}_{\phi_{1,1}}$, $\text{V}_{\phi_{1,2}}$, $\text{D}_{\phi_2}$ and $\text{G}_{\phi_3}$ correspondingly. This form of the neural network(or more precisely, the combination of the neural networks) allows learning the dynamic governed by the Equation \eqref{eq: port-Hamilt} from the data. 

Let us introduce the notions $\text{q} = \nabla F(\theta_{t})$ and $\text{p} = \text{M} \dot{\text{q}}$, then the update propose by our model is the following:
\begin{equation*}
    f_{\phi}(\nabla F(\theta_{t}), \dot{\nabla F(\theta_{t})}) = W[\text{q}, \dot{\text{q}}]^T,
\end{equation*}
where $\dot{\text{q}}$ is obtained from the Equation \eqref{eq: port-Hamilt}. The intuition behind this is explained below. The Equation \eqref{eq: port-Hamilt} takes as an input $\text{q} = \nabla F(\theta_{t})$ and $\text{p} = \text{M} \dot{\text{q}}$, the first is calculated by using usual backpropagation at each iteration, the second is calculated by taking $\dot{\text{q}}$ from the output of Equation \eqref{eq: port-Hamilt} at the previous iteration. We call our model the Hamiltonian Neural Network (HNN). In \cite{andrychowicz2016learning} authors use LSTM model \cite{sak2014long} for $f_{\phi}$.

Let us take a closer look at the structure of our model. The scheme of our model is presented in the Fig. \ref{HNN procedure}. The blue and red blocks are the model's inputs and outputs correspondingly. The light gray blocks have the same denotation as in the Equation \eqref{eq: port-Hamilt} and are learnable. It takes as an input the vector of two stacked components at time $t$ $[\nabla F(\theta_{t}), \dot{(\nabla F(\theta_{t}))}]^T$: the gradient of the objective $\nabla F(\theta_{t})$ and its time derivative $\dot{(\nabla F(\theta_{t}))}$ (blue blocks in the Fig. \ref{HNN procedure}). As a result, physical part of the model returns the vector $[\dot{(\nabla F(\theta_{t}))}, \ddot{(\nabla F(\theta_{t}))}]^T$ of stacked derivatives of the input components at time $t$ (green blocks in the Fig. \ref{HNN procedure}). By taking the product of the first output component $\dot{\nabla F(\theta_{t})}$ and a small constant $\Delta t$, we can obtain a rough approximation of the corrected gradient at the next time step $\nabla F(\theta_{t+1})$:\[\nabla F(\theta_{t+1}) \approx \nabla F(\theta_{t}) + \dot{\nabla F(\theta_{t})} \Delta t.\]
Such a scheme allows us to correct the gradient to make it obey the learned dynamic. 
As we do not know the optimal value of the constant $\Delta t$, we propose to use a linear layer without bias and with matrix $W \in \mathcal{R}^{2 \times 1}$ that allows us to learn how to combine the terms $\nabla F(\theta_{t})$ and $\dot{\nabla F(\theta_{t})}$ in the optimal way (top orange block in the Fig. \ref{HNN procedure}):
\[\nabla\Tilde{F}(\theta_{t+1}) \approx W_{1} \nabla F(\theta_{t}) + W_{2}\dot{\nabla F(\theta_{t})},\]
where approximation is true up to the multiplication by some constant (can be leveled by the smaller/bigger learning rate step). The same procedure we apply to approximate $\dot{\nabla F(\theta_{t+1})}$ with another linear layer with the inputs $\dot{\nabla F(\theta_{t})}$ and $\ddot{\nabla F(\theta_{t})}$ (bottom orange block in the Fig. \ref{HNN procedure}). Remind the reader that we need the component $\dot{\nabla F(\theta_{t+1})}$ as an input into our model at the next iteration. For simplicity we assume that inertia matrix $\text{M}$ does not depend on $\text{q}$. This let us to do the direct approximation of the partial derivative: $$\frac{\partial H(\text{q},\text{p})}{\partial \text{q}} \approx \frac{\partial H_{\phi_1}(\text{q},\text{p})}{\partial \text{q}} = \text{V}_{\phi_{1,2}}(\text{q}).$$ Thus, for the Equation \eqref{eq: port-Hamilt} there is no reason to compute Hamiltonian $H(\text{q},\text{p})$ as $$\frac{\partial H(\text{q},\text{p})}{\partial \text{p}} \approx \text{M}^{-1} p = \text{M}^{-1}_{\phi_{1,1}} p,$$ so we know all the terms dependant on the Hamiltonian. Moreover, there is no reason to compute inertia matrix itself, but only its inverse. Hence we approximate inverse of the inertia matrix directly to escape the computational issues: $\text{M}^{-1} \approx \text{M}^{-1}_{\phi_{1,1}} = L_{\phi_{1,1}}^TL_{\phi_{1,1}},$ this form is due to the fact that $\text{M}$ is the positive semi-definite matrix.

There were several attempts to learn the optimizer using the recurrent neural networks, specifically LSTM \cite{ravi2016optimization,younger2001meta}. To the best of our knowledge, it is a first attempt to learn an optimizer with the hidden physical structure.

\section{Experiments}

In this section, we experimentally compare the results of the proposed model against the LSTM model from prior work \cite{andrychowicz2016learning} and standard optimization methods used in deep learning such as ADAM \cite{kingma2014adam}, RMSprop \cite{hinton2012neural}, SGD \cite{ruder2016overview}, and NAG \cite{nesterov1983method}. For the standard optimizers and LSTM-based one, we repeat experiment settings reported in \cite{andrychowicz2016learning} where each of these optimizers learning rate was tuned. 

We take an update step as an output of the neural network. Similar to \cite{andrychowicz2016learning}, to train this neural network we use an objective function for the training that depends on the part of the trajectory of optimization, for some horizon $T$:
\begin{equation} \label{eq: objective}
    \mathcal{L}(\phi) = \sum_{t=1}^T F(\theta_t),
\end{equation}
where $\theta_{t+1} = \theta_t + f_{\phi}(\nabla F(\theta_{t}))$.

We consider two experiment settings: minimization of the quadratic function and the optimization of the base network on the MNIST dataset. The optimization of the HNN is done using ADAM with the learning rate $0.01$ for both experiment settings, no weight decay is used. We pick the best parameters for our model according to the validation loss which is calculated during the training after each epoch. Finally, in each experiment, we report the average performance on a number of freshly sampled test problems.

\subsection{Quadratic functions}
In this section, for the optimization we consider 10-dimensional quadratic functions of the form: 
\begin{equation*}
    F(\theta) = ||W \theta - y||^2_2,
\end{equation*}
where $W \in \mathcal{R}^{10 \times 10}$ and $y \in \mathcal{R}^{10}$ that is drawn IID from the Gaussian distribution.
Each function is optimized using objective \eqref{eq: objective} for 100 steps with the horizon parameters $T=20$ and $T=15$ for LSTM and HNN correspondingly.

\begin{figure}
\begin{center}
\centerline{\includegraphics[width=0.9\columnwidth]{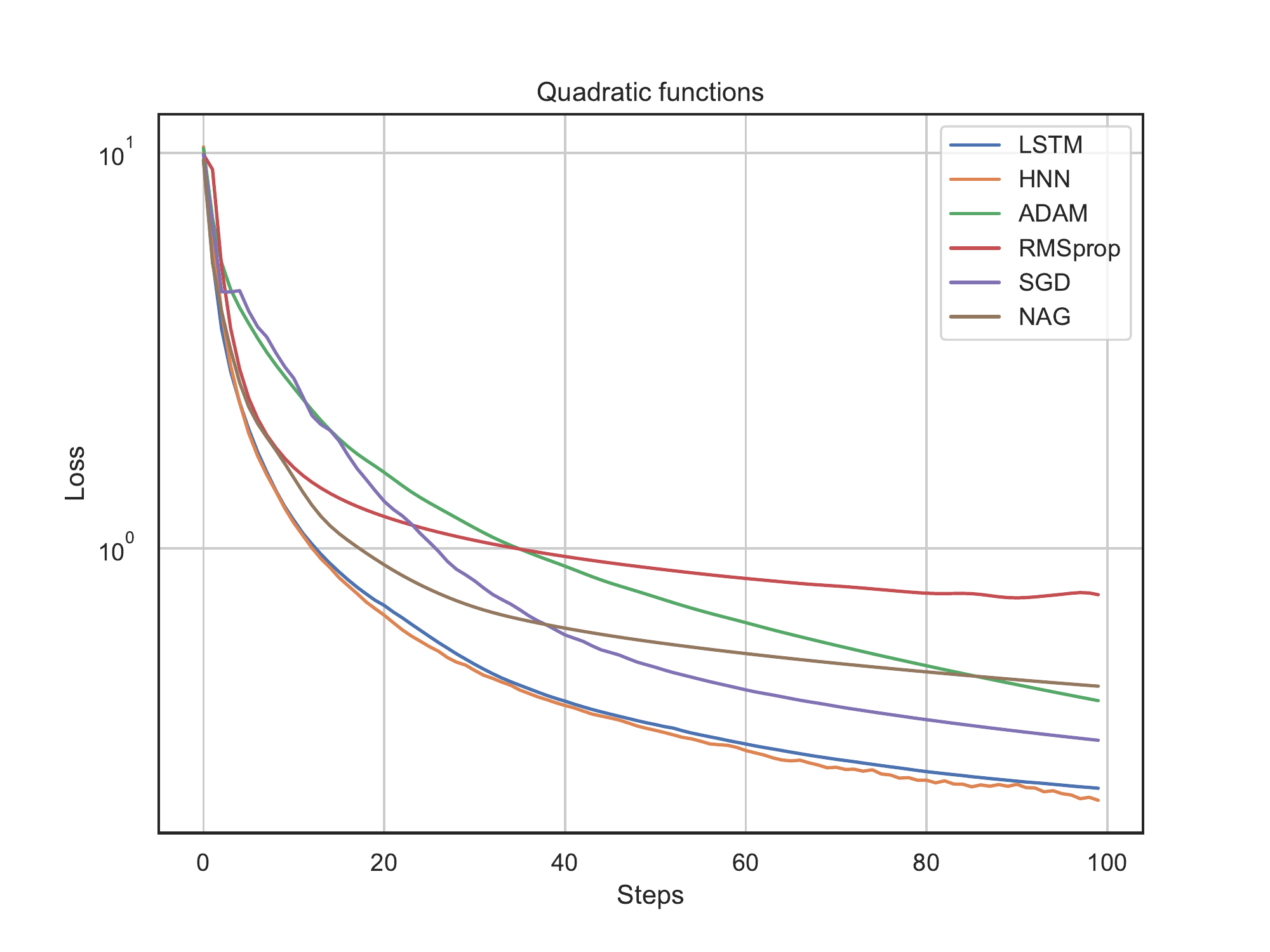}}
\caption{Training curves for the quadratic function}
\label{fig: quadratic}
\end{center}
\end{figure}

The results are presented in the Tab. \ref{tab:quadratic}. In the Fig. \ref{fig: quadratic} learning curves for different optimizers are presented. Each curve corresponds to the average performance over 100 test functions. One can see that both learned optimizers outperform the standard optimizers and LSTM slightly underperforms HNN which has significantly fewer parameters. 

\subsection{MNIST}
In this part, we train the optimizer model to optimize a base network on the MNIST dataset. The objective function $F(\theta)$ is the cross-entropy for the base network that is MLP with 20 units and a sigmoid activation function. The optimization was run for 100 steps with the horizon parameter $T= 20$ and $T = 15$ for the LSTM and HNN correspondingly. We evaluate each optimization approach over 100 test functions on the two base networks: with the sigmoid and ReLU activation functions. Finally, we present the average results. The source of variability between different runs is the initial value of the base model parameters $\theta_0$ and the order of batches of data. 

\begin{figure}
\begin{center}
\centerline{\includegraphics[width=0.9\columnwidth]{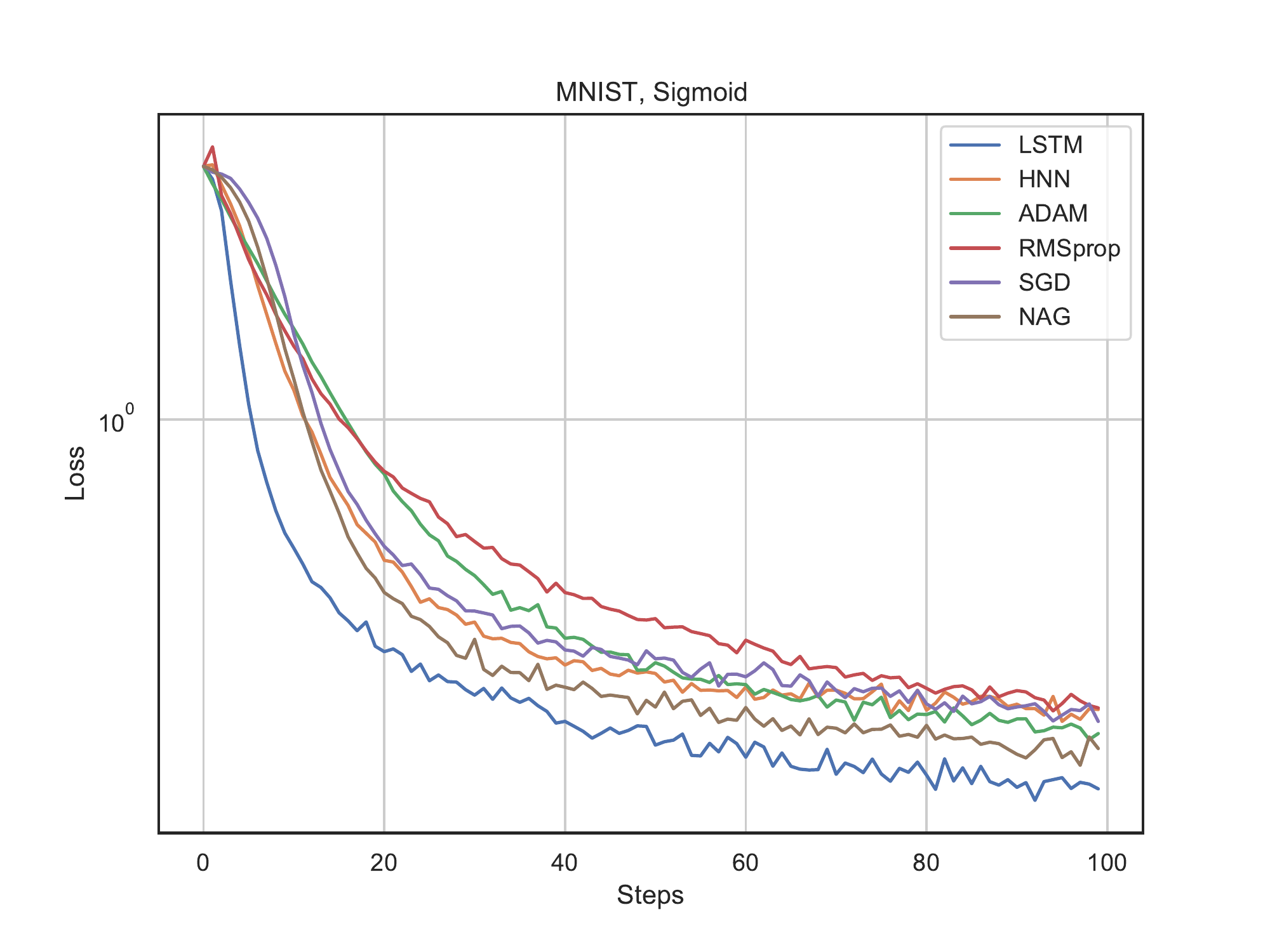}}
\caption{Training curves for MNIST with the Sigmoid activation}
\label{fig:mnist_sigm}
\end{center}
\end{figure}

\begin{figure}
\begin{center}
\centerline{\includegraphics[width=0.9\columnwidth]{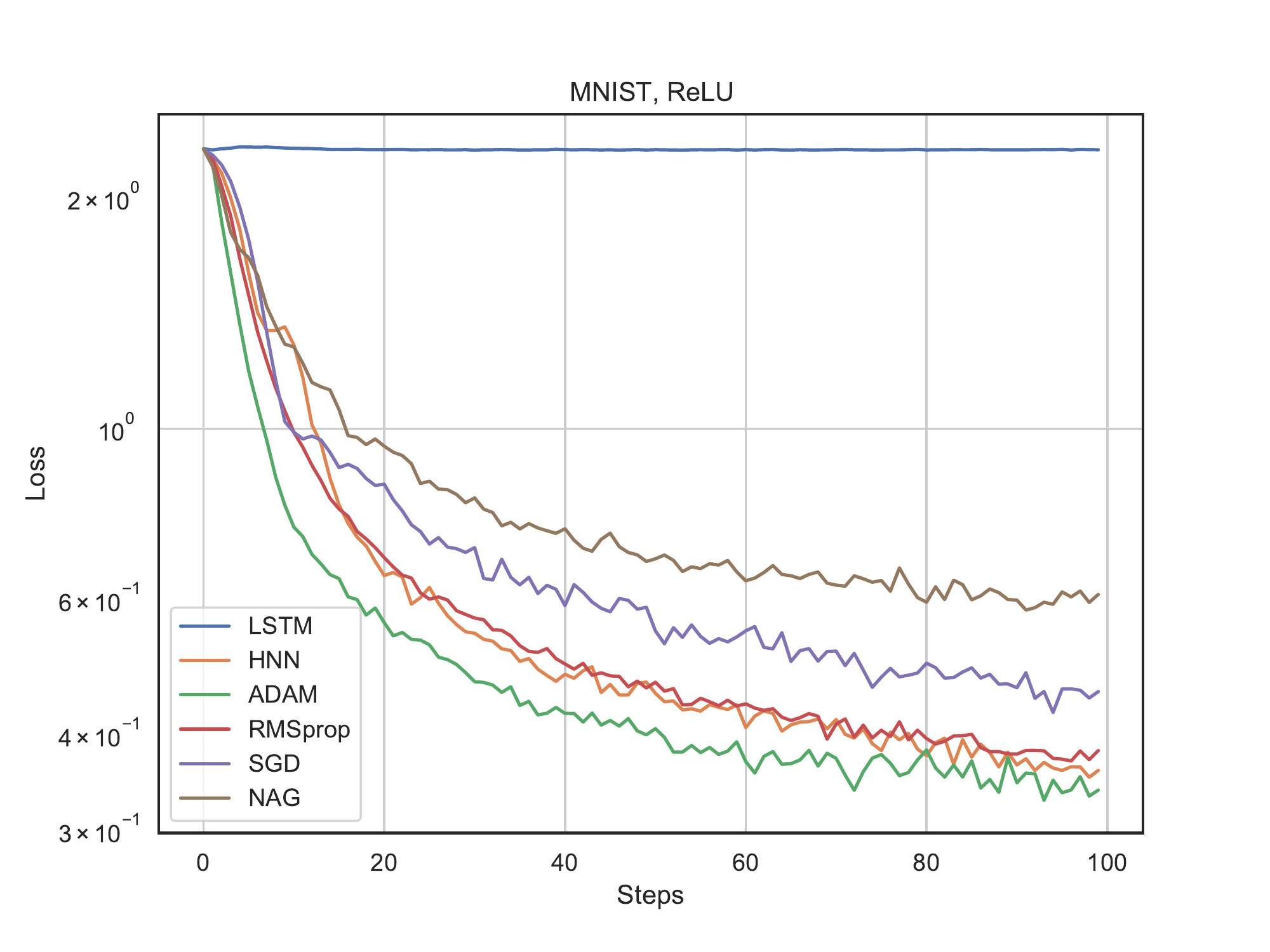}}
\caption{Training curves for MNIST with the ReLU activation}
\label{fig:mnist_relu}
\end{center}
\end{figure}

The comparison of the results for the sigmoid and ReLU are presented in the Tab. \ref{tab:mnist_sigmoid} and Tab. \ref{tab:mnist_ReLU} correspondingly. Averaged over 100 runs learning curves for the base network using different optimizers are shown in the Fig. \ref{fig:mnist_sigm} and Fig. \ref{fig:mnist_relu}. The results for sigmoid show that HNN comparable or slightly worse than other methods which we relate to the shallowness of our model and, as a result, weak expressive power. At the same time, we see from the results for ReLU that our model has better transfer properties than LSTM and produces comparable results to other standard methods.

\section{Discussion}
First, this work has started from the idea to apply Neural Ordinary Differential Equations (Neural ODE) \cite{chen2019neural} to the given problem. That is, the Equation \eqref{eq: port-Hamilt} is the base of Neural ODE, which takes as an input parameter $\text{q} = \theta_t$ and returns $\theta_{t+1}$ after the integration. Another idea is to do the same procedure for the gradient of the objective function at point $\theta_t$, that is, $\text{q} = \nabla F(\theta_t)$ and after integration, we obtain the 'corrected' version of the gradient $\nabla F(\theta_{t+1})$ according to the learned dynamic. Experiments with the described approaches are quite time-consuming (due to the integration part in the Neural ODE) and the results are not promising. Due to simplicity and superiority in terms of performance over above discussed approaches we stick to the proposed in this paper method: without the use of ODE and with $\text{q} = \nabla F(\theta_t)$ we obtain not the $\nabla F(\theta_{t+1})$, but $\dot{\nabla F(\theta_{t+1})}$ that after multiplying by some small constant can be seen as a rough approximation of the change in the gradient according to the learned dynamic.

\section{Conclusion}
Due to specificity, trained for the given problem optimizers produce better or comparable results with respect to standard optimizers. From our experiments, one can see that the learned neural optimizer with the hidden physical structure produces comparable performance against the proposed in the prior work LSTM optimizer and widely used gradient methods while having much fewer parameters. Moreover, it has better generalization than the LSTM model because it is not over-parametrized and thus not overfitted on MNIST with the sigmoid activation. This shows that gradients can be learned by HNN and we can benefit from the induced implicit bias at least in the simple optimization of a quadratic function. 


\bibliographystyle{ieeetr}
\bibliography{biblio}
\newpage
\appendix

\setcounter{figure}{0}  
\setcounter{table}{0} 
\renewcommand\thefigure{A\arabic{figure}}  
\renewcommand\thetable{A\arabic{table}}

\begin{figure}[!htbp]
    \centering
    \includegraphics[scale = 0.14]{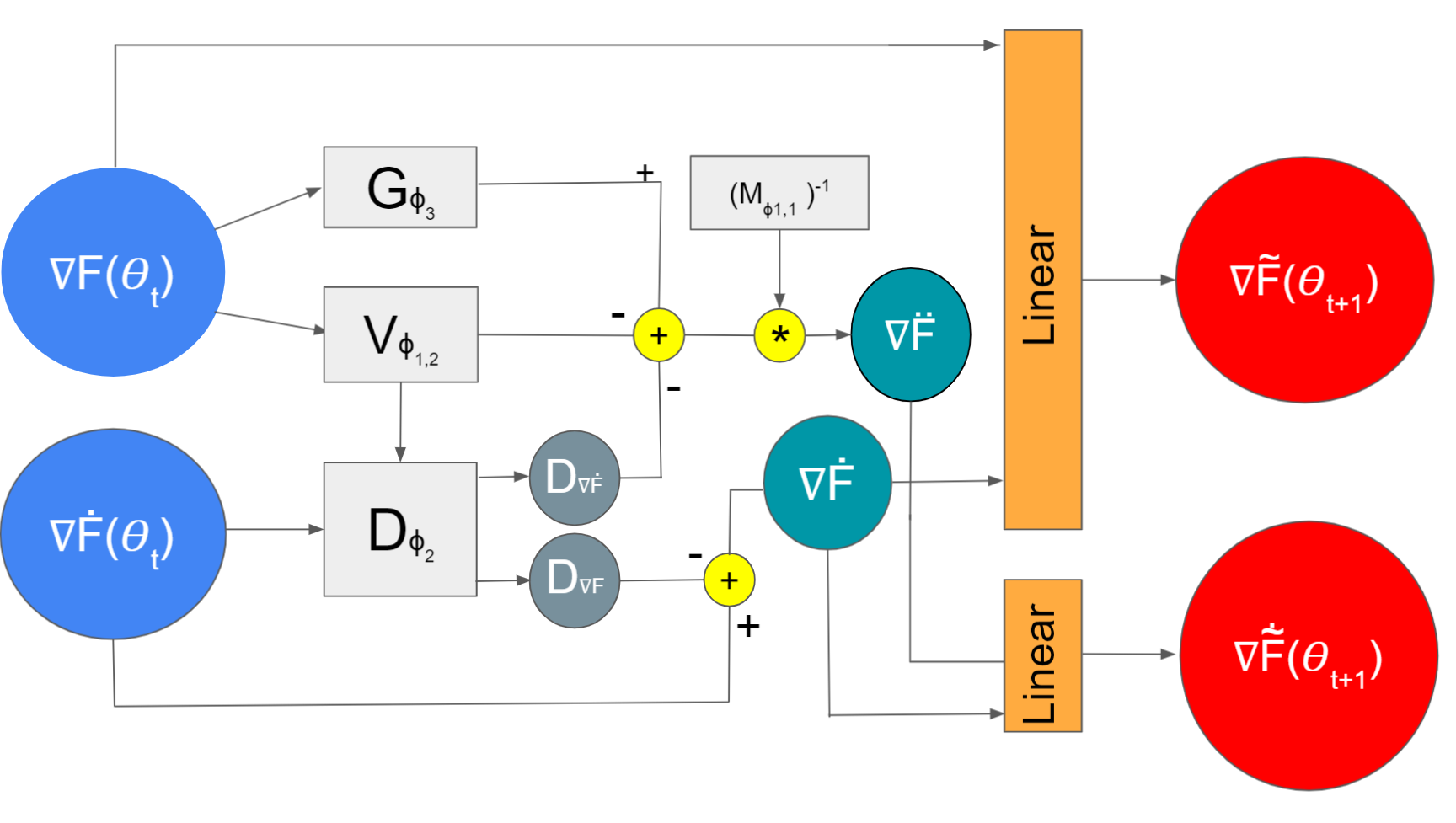}
    \caption{The computation graph of HNN.}
    \label{HNN procedure}
\end{figure}

\begin{table}[h!]
\begin{center}
\begin{small}
\begin{sc}
\begin{tabular}{lccc}
\toprule
Method & \# Params & Best Loss ($\downarrow$) \\
\midrule
ADAM & - & 0.41 \\
RMSprop & - & 0.76 \\
SGD & - & 0.33 \\
NAG & - & 0.45 \\
LSTM & 5221 & 0.25 \\
HNN & 19  &  \textbf{0.23} \\
\bottomrule
\end{tabular}
\end{sc}
\end{small}
\end{center}
\caption{Results on the quadratic function.}
\label{tab:quadratic}
\end{table}

\begin{table}[h!]
\begin{center}
\begin{small}
\begin{sc}
\begin{tabular}{lccc}
\toprule
Method & \# Params & Best Loss ($\downarrow$) \\
\midrule
ADAM & - &  0.35\\
RMSprop & - &  0.38\\
SGD & - &  0.36\\
NAG & - &   0.33\\
LSTM & 5301 & \textbf{0.29} \\
HNN & 55 & 0.38\\
\bottomrule
\end{tabular}
\end{sc}
\end{small}
\end{center}
\caption{Results on MNIST dataset, Sigmoid.}
\label{tab:mnist_sigmoid}
\end{table}

\begin{table}[h!]
\begin{center}
\begin{small}
\begin{sc}
\begin{tabular}{lccc}
\toprule
Method & \# Params & Best Loss ($\downarrow$) \\
\midrule
ADAM & - & \textbf{0.34} \\
RMSprop & - & 0.38 \\
SGD & - & 0.45 \\
NAG & - & 0.61 \\
LSTM & 5301 & 2.31 \\
HNN & 55 & 0.36\\
\bottomrule
\end{tabular}
\end{sc}
\end{small}
\end{center}
\caption{Results on MNIST dataset, ReLU.}
\label{tab:mnist_ReLU}
\end{table}
\end{document}